\documentclass{article}

\usepackage[preprint]{neurips_2023}

\usepackage{algorithm, algorithmicx, algpseudocode}
\usepackage{amsmath, amsfonts, amssymb}
\usepackage{array}
\usepackage{booktabs}
\usepackage{collcell}
\usepackage{datatool}
\usepackage{enumitem}
\usepackage{environ}
\usepackage{etoolbox}
\usepackage{flushend}
\usepackage{graphicx}
\usepackage{lipsum}
\usepackage[nopatch=eqnum]{microtype}
\usepackage{multirow}
\usepackage{nicefrac}
\usepackage{printlen}
\usepackage{soul}
\usepackage{svg}
\usepackage{url}
\usepackage{wrapfig}
\usepackage{xcolor,colortbl}
\usepackage{xstring}

\definecolor{Gray}{gray}{0.85}
\definecolor{cite}{HTML}{53769A}
\definecolor{ref}{HTML}{379030}
\definecolor{lightcornflowerblue}{rgb}{0.6, 0.81, 0.93}
\definecolor{lightkhaki}{rgb}{0.94, 0.9, 0.55}
\definecolor{lightmauve}{rgb}{0.86, 0.82, 1.0}
\definecolor{lightgreen}{rgb}{0.56, 0.93, 0.56}

\definecolor{royalpurple}{RGB}{207,199,216}
\definecolor{forestgreen}{RGB}{202,225,204}
\definecolor{red}{RGB}{247,207,185}

\definecolor{PatternA}{RGB}{180, 22,  0   }
\definecolor{PatternC}{RGB}{23,  77,  127 }
\definecolor{PatternB}{RGB}{55,  144, 48  }

\usepackage[breaklinks,colorlinks=true, citecolor=cite, linkcolor=ref]{hyperref}

\soulregister\Hl{7}
\soulregister\ref7
\soulregister\cite7
\soulregister\pageref7

\newcommand{\Hl}[2][\empty]{%
\ifx#1\empty
\else
\sethlcolor{#1}%
\fi
\hl{#2}}

\newcommand\citecolor[1]{%
  \expandafter\newcommand\csname citecolor-#1\endcsname{}%
}

\newcolumntype{H}{>{\setbox0=\hbox\bgroup}c<{\egroup}@{}}
\algrenewcommand{\algorithmiccomment}[1]{\hfill$\blacktriangleright$ #1}

\newcolumntype{X}{>{\columncolor{lightcornflowerblue}}c}
\newcolumntype{Y}{>{\columncolor{lightkhaki}}c}
\newcolumntype{Z}{>{\columncolor{lightmauve}}c}
\newcolumntype{P}{>{\columncolor{lightgreen}}c}

\newcommand{\com}[1]{\iffalse~#1~\fi}%
\newcommand{\x}[1]{{\underline{#1}}}%

\newcommand{\noimage}{%
  \setlength{\fboxsep}{-\fboxrule}%
  \fbox{\phantom{\rule{100pt}{100pt}}File missing\phantom{\rule{100pt}{100pt}}}
}
\let\includegraphicsoriginal\includegraphics%
\renewcommand{\includegraphics}[2][width=\textwidth]{%
    \IfFileExists{#2}{\includegraphicsoriginal[#1]{#2}}{\noimage}
}

\emergencystretch=\maxdimen%
\everypar{\looseness=-1 }

\newcounter{descriptcount}

\newcounter{CurrentRow}
\newcounter{CurrentColumn}
\newtoggle{DoneWithFirstRow}

\newcommand*{\FirstColumn}[1]{%
    \IfEq{\arabic{CurrentColumn}}{0}{%
        \global\togglefalse{DoneWithFirstRow}%
        \setcounter{CurrentRow}{1}
    }{%
        \global\toggletrue{DoneWithFirstRow}%
        \stepcounter{CurrentRow}%
    }%
    \setcounter{CurrentColumn}{0}%
    \NewData{#1}%
}
\newcommand*{\NewData}[1]{%
    \dtlexpandnewvalue%
    \stepcounter{CurrentColumn}%
    \iftoggle{DoneWithFirstRow}{%
        \dtlgetrow{TransposedTabularDB}{\arabic{CurrentColumn}}%
        \dtlappendentrytocurrentrow{\Alph{CurrentRow}}{#1}%
        \dtlrecombine%
    }{%
        \DTLnewrow{TransposedTabularDB}%
        \DTLnewdbentry{TransposedTabularDB}{\Alph{CurrentRow}}{#1}%
    }%
}%

\newcolumntype{+}{>{\global\let\currentrowstyle\relax}}
\newcolumntype{^}{>{\currentrowstyle}}
\newcolumntype{F}{>{\collectcell\FirstColumn}c<{\endcollectcell}}
\newcolumntype{C}{>{\collectcell\NewData}{c}<{\endcollectcell}}

\newtoggle{EncounteredDataRow}

\newsavebox{\TempBox}
\DTLnewdb{TransposedTabularDB}

\NewEnviron{Ttabular}[1]{%
    \setcounter{CurrentColumn}{0}%
    \global\togglefalse{EncounteredDataRow}%
    \savebox{\TempBox}{%
        \begin{tabular}{FCCCCCC}
            \BODY%
        \end{tabular}%
    }%
    \begin{tabular}{#1}\toprule%
    \DTLforeach*{TransposedTabularDB}{\Aa=A, \Ba=B, \Ca=C}{%
      \DTLiffirstrow{}{\\\midrule}%
        \Aa&\Ba&\Ca%
    }\\\bottomrule%
    \end{tabular}%
}%

\def\Tableref#1{Table~\ref{#1}}

\def\eqref#1{equation~\ref{#1}}

\usepackage{mathabx,graphicx}

\def\Figref#1{Figure~\ref{#1}}

\def\1{\bm{1}}
\newcommand{\train}{\mathcal{D}}

\newcommand{\E}{\mathbb{E}}
\newcommand{\Ls}{\mathcal{L}}
\newcommand{\R}{\mathbb{R}}

\DeclareMathOperator*{\argmax}{arg\,max}
\DeclareMathOperator*{\argmin}{arg\,min}

\newenvironment{aequation}
{\begin{equation*} \begin{aligned}}
{\end{aligned} \end{equation*}}

\DeclareMathAlphabet{\mathsfit}{\encodingdefault}{\sfdefault}{m}{sl}
\SetMathAlphabet{\mathsfit}{bold}{\encodingdefault}{\sfdefault}{bx}{n}

\def\sP{{\mathbb{P}}}

\title{Towards Improving Robustness Against Common Corruptions using Mixture of Class Specific Experts}

\author{%
Shashank Kotyan \quad Danilo Vasconcellos Vargas \\
Laboratory of Intelligent Systems \\
Kyushu Univeristy, Fukuoka, Japan \\
}

\begin{document}

\maketitle

\begin{abstract}
Neural networks have demonstrated significant accuracy across various domains, yet their vulnerability to subtle input alterations remains a persistent challenge.
Conventional methods like data augmentation, while effective to some extent, fall short in addressing unforeseen corruptions, limiting the adaptability of neural networks in real-world scenarios.
In response, this paper introduces a novel paradigm known as the Mixture of Class-Specific Expert Architecture.
The approach involves disentangling feature learning for individual classes, offering a nuanced enhancement in scalability and overall performance.
By training dedicated network segments for each class and subsequently aggregating their outputs, the proposed architecture aims to mitigate vulnerabilities associated with common neural network structures.
The study underscores the importance of comprehensive evaluation methodologies, advocating for the incorporation of benchmarks like the common corruptions benchmark.
This inclusion provides nuanced insights into the vulnerabilities of neural networks, especially concerning their generalization capabilities and robustness to unforeseen distortions.
The research aligns with the broader objective of advancing the development of highly robust learning systems capable of nuanced reasoning across diverse and challenging real-world scenarios.
Through this contribution, the paper aims to foster a deeper understanding of neural network limitations and proposes a practical approach to enhance their resilience in the face of evolving and unpredictable conditions.

\end{abstract}

\section{Introduction}

We've leveraged the capabilities of neural networks to achieve high precision across a variety of tasks applied in diverse fields such as natural language processing and computer vision.
Many of these tasks would be unfeasible without the assistance of neural networks.
Nevertheless, the existing neural networks~\citep{krizhevsky2012ImageNet} lack the resilience observed in the human visual system~\citep{recht2018CIFAR10, recht2019ImageNet, azulay2019Why}.
Neural networks tend to make mistakes when confronted with minor modifications in input samples~\citep{szegedy2014Intriguing, carlini2016Defensive, carlini2017Adversarial}, in contrast to the robustness of the human vision system against slight alterations in images~\citep{dodge2017Study}.
Humans effortlessly navigate through various corruptions and distortions in images, and even abstract changes in structure and style do not disrupt their visual perception.

Earlier research suggests enhancing training data by introducing various distortions to improve overall resilience.
However, obtaining training data for rare conditions is challenging, and it's impossible to foresee all potential conditions that might occur in real-world situations.
Recent findings indicate that neural networks struggle to generalize to newly introduced distortion types, even when trained on a diverse set of other distortions, as shown by~\cite{geirhos2018Generalisation}.
In a broader context, neural networks often face difficulty in extending their performance beyond the domain or distribution of the training data.
As a result, the generation of adversarial examples, as demonstrated by~\cite{szegedy2014Intriguing}, becomes possible.

Adversarial examples represent the most severe attempts to exploit the vulnerabilities of neural networks, characterized by a minimal domain shift that is imperceptible to humans yet sufficient to deceive the neural network.
Therefore, analyzing with adversarial examples involves considering the worst-case scenario for a neural network rather than the more common average-case scenarios encountered in real-world situations.
To address the prevalent but less extreme issue of noticeable image distortions, such as blurriness, noise, or natural distortions like snow, one can utilize the analysis proposed by~\cite{hendrycks2019Benchmarking} through their suggested common corruptions benchmarks.
This allows for an understanding of how performance degrades in situations closely resembling real-world conditions.

Research by \cite{kotyan2022Transferability} suggests that current deep neural networks generalise poorly over the unknown classes which is also related to their robustness.
This suggests that the network overfit on the trained classes such that they generalise poorly on the unknown classes.
Another research by \cite{mathieu2019Disentangling} suggests that disentangling features in a representation is crucial for understanding and interpreting the underlying factors of variation in data.
Thus, building on the past researches by \cite{kotyan2022Transferability,mathieu2019Disentangling} we propose to learn the disentagled features by making a part of the network focus on a particular feature in order to better understand the underlying variations and learn more robustly.

\textbf{Contributions.}
\begin{description}[style=sameline, leftmargin=*]

    \item[Mixture of Class-Specific Expert Architecture:]
    We suggest employing a combination of class-specific expert architectures to enhance overall robustness by separating the feature learning process for distinct classes within a dataset.
    Our approach involves training a segment of the network exclusively focused on recognizing features specific to a single class.
    Subsequently, we aggregate the outcomes from all individual class-specific experts to determine the image classification.
    This disentanglement of feature learning not only enhances performance but also improves robustness, making it applicable to datasets with a large number of images.

\end{description}

\section{Related Works}

\textbf{Challenges in the field of Adversarial Machine Learning}.
Adversarial images undergo carefully crafted distortions aiming to confuse classifiers, occasionally deceiving black-box classifiers \cite{kurakin2017Adversarial}.
Algorithms have been developed to identify minimal additive distortions in RGB space that can confuse classifiers \citep{carlini2018GroundTruth}.
Neural networks exhibit peculiar behavior for nearly identical images, showing high confidence when faced with textures and random noise, revealing vulnerabilities exploited by adversarial attacks \citep{szegedy2014Intriguing,nguyen2015Deep}.
The feasibility of universal adversarial perturbations, capable of deceiving a neural network across most samples, has been demonstrated (Moosavi-Dezfooli et al., 2017).
The introduction of patches into an image can lead to misclassification by neural networks \citep{brown2018Adversarial}, and extreme attacks have proven effective, causing misclassification with a single-pixel change \citep{su2019One, kotyan2022Adversarial}.
To counter adversarial attacks, \cite{goodfellow2015Explaining} proposed the Fast Gradient Sign Method (FGSM), perturbing a target sample to its gradient direction to increase its loss and using the generated samples to train the model for improved robustness.
Subsequent works introduced iterative variants of the gradient attack with enhanced adversarial learning frameworks \citep{madry2018Deep, moosavi-dezfooli2016DeepFool, kurakin2017Adversarial, carlini2017Evaluatinga}.

\textbf{Challenges in evaluating natural corruptions, perturbations and distortions}.
Various studies highlight the susceptibility of neural networks to common corruptions, with \cite{hosseini2017Google} disrupting Google’s Cloud Vision API using impulse noise, and \cite{dodge2016Understanding} demonstrating the sensitivity of Convolutional Neural Networks to blur and Gaussian noise in image recognition.
Another study by \cite{dodge2017Study} employs Gaussian noise and blur to underscore the superior robustness of human vision compared to neural networks, even after specific fine-tuning for these distortions.
\cite{geirhos2018Generalisation} find that neural network performance declines more rapidly than human performance in recognizing corrupted images, and fine-tuning on specific corruptions does not lead to generalization.
\cite{hendrycks2019Benchmarking} introduce a benchmark of $19$ common corruptions to assess model robustness, and various approaches, such as preprocessing to eliminate corruptions or data augmentation by including corrupted data in training, are explored to address performance degradation.
Techniques like matching softmax distributions or using a mixture of corruption-specific experts, as suggested by \cite{zheng2016Improving} and \cite{dodge2017Quality}, aim to mitigate issues such as under-fitting when fine-tuning on noisy images.
\cite{geirhos2018ImageNettrained} train recognition models on a stylized version of the ImageNet dataset, reporting increased robustness against different corruptions due to a stronger bias towards object shape over textures.
Additionally, \cite{hendrycks2019Benchmarking} present methods like Histogram Equalization~\citep{delatorre2005Histogram, harvilla2012Histogrambased}, Multi-scale Networks~\citep{ke2017Multigrid, huang2018MultiScale}, Adversarial Logit Pairing~\citep{kannan2018Adversarial}, Feature Aggregating~\citep{xie2017Aggregated}, and Larger Networks to enhance performance on their corruption benchmark.
These efforts collectively seek to improve neural network resilience to a diverse set of common corruptions.

\textbf{Challenges in disentanglement}.
Disentanglement in machine learning, refers to achieving independence among features in learned representations.
This is acheived as features are disentangled each dimension of the representation captures a specific and independent aspect of the data, promoting transparency and interpretability \citep{bengio2013Representation,eastwood2018Framework,higgins2018Definition}.
This concept has roots in traditional methods like Independent Component Analysis and extends to modern deep learning approaches enableing the model to learn meaningful and separable representations of complex data  \citep{schmidhuber1992Learning, yang1997Adaptive, hyvarinen2000Independent, reed2014Learning, cheung2015Discovering, chen2016InfoGAN, makhzani2016Adversarial, mathieu2016Disentangling, achille2018Emergencea, hjelm2019Learning, mathieu2019Disentangling}.
The primary goal is to create interpretable representations by capturing meaningful and independent dimensions in the data.
Disentanglement is crucial for identifying true generative factors within the data, enhancing transparency, and facilitating tasks like classification and generation \citep{higgins2016betaVAE, n2017Learning, alemi2018Fixing, chen2018Isolating, kim2018Disentangling, esmaeili2019Structured}.
Challenges arise in capturing the true generative factors for complex datasets, necessitating richly structured dependencies between latent dimensions \citep{eastwood2018Framework}.
Despite challenges, the development of decomposition frameworks aims to overcome limitations in existing disentanglement approaches, highlighting the need for structured dependencies in more complex datasets \citep{johnson2016Composing, n2017Learning,bouchacourt2018MultiLevel,esmaeili2019Structured}.

\section{Formal Mathematical Formulations}

Let us suppose that for the image classification problem, we consider a classifier
$f_\theta : x \rightarrow y$
trained on samples from distribution $D$ such that
$x \in \mathbb{R}^{m \times n \times c}$
be the  image which is to be classified.
Here $m, n$ is the width and the height of the image, and $c$ is the number of colour channels.

A neural network is composed of several neural layers linked together.
Each neural layer is composed of a set of perceptrons (artificial neurons).
Each perceptron maps a set of inputs to output values with an activation function.
Thus, function of the neural network (formed by a chain) can be defined as:
\begin{aequation}
f_\Theta(x) = g_{\theta_k}^{(k)}( \ldots g_{\theta_2}^{(2)}(g_{\theta_1}^{(1)}(x)))
\end{aequation}
where $g_{\theta_i}^{(i)}$ is the function of the $i^{\text{th}}$ layer of the network, and $ i = 1,2,3 \ldots k$ such that $k$ is the last layer of the neural network.
$\theta_i$ is the parameter of the $i^{\text{th}}$ layer which is optimised and consequently $\Theta = \{ \theta_1, \theta_2, \theta_3 \ldots \theta_k \}$ is set of all parameters of the neural network $f$ which is optimised.
In the image classification problem,
$f(x) \in \mathbb{R}^{N}$
is the probabilities (confidence) for all the available $N$ classes.
Most classifiers are judged by their performance (often measured in accuracy) on test queries drawn from $\train$, i.e.,
\begin{aequation}
\mathbb{P}_{(x, y) \sim \train} ~~\underset{y}{\argmax} & & f(x) = y~~.
\end{aequation}

\subsection{Adversarial Perturbations}

Let us define adversarial samples $\hat{x}$ as:
\begin{aequation}
& \hat{x} = x + \epsilon_{x}
& \{ \hat{x} \in \R^{m \times n \times 3} \mid \underset{y}{\argmax} ~~ f(x) \ne \underset{\hat{y}}{\argmax}~~f(\hat{x})  \}
\end{aequation}
in which
$\epsilon_{x} \in \R^{m \times n \times c}$
is the perturbation added to the input.
Here, $y$ and $\hat{y}$ are the respective classfications for $x$ and $\hat{x}$.

Making use of the definition of adversarial samples, adversarial robustness can be formally defined as the following optimization problem for untargeted black-box attacks:
\begin{aequation}
& \underset{\epsilon_{x}}{\text{minimize}} & & f(\hat{x})_y = f(x+\epsilon_{x})_y
& \text{subject to} & & \Vert \epsilon_{x} \Vert_p \leq \delta
\end{aequation}
Similarly optimization problem for the targeted black-box attacks can be defined as:
\begin{aequation}
& \underset{\epsilon_{x}}{\text{maximize}} & & f(\hat{x})_{\tilde{y}} = f(x+\epsilon_{x})_{\tilde{y}}
& \text{subject to} & & \Vert \epsilon_{x} \Vert_p \leq \delta
\end{aequation}
where $f()_c$ is the soft-label for the class $c$, and $y$ is the true class of sample $x$, whereas $\tilde{y}$ is target class for the the sample $x$.
$p$ is the constraint on $\epsilon_{x}$ and $\delta$ is the threshold value for the constraint.
Thus, adversarial robustness can be formulated as,
\begin{aequation}
\underset{\Vert \epsilon_{x} \Vert_p ~\leq~ \delta}{\text{minimize}} & & \sP_{(x, y) \sim \train} ~~\underset{\hat{y}}{\argmax} & & f(\hat{x}) = y
\end{aequation}
Given such a dataset $\train$ and a model $f$, adversarial attacks aim towards finding the worst-case examples nearby by searching for the perturbation $ \epsilon $, which maximizes the loss.
We can define one such adversarial attack which tries to find perturbation within a certain radius from the sample (e.g., norm balls) as follows:
\begin{aequation}
\hat{x}^{i+1} = \Pi_{\Vert \epsilon_{x} \Vert_p ~\leq~ \delta}(\hat{x}^{i} + \alpha \cdot \text{sign}(\nabla_{\hat{x}^{i}} \Ls_{\text{CE}, \theta}(f(\hat{x}^{i}), y)))
\end{aequation}
where $\Vert \epsilon_{x} \Vert_p ~\leq~ th$ is the norm-ball around $x$ with radius $th$, and $\Pi$ is the projection function for norm-ball.
The $\alpha$ is the step-size of the attacks whereas sign(·) returns the sign of the vector and $\nabla_{p}(q)$ is the gradient of $p$ w.r.t $q$.
Further, $\mathcal{L}_{\text{CE}}$ is the cross-entropy loss for supervised training, and $i$ is the number of attack iterations.
This formulation generalizes across different types of gradient attacks.
For example, Projected Gradient Descent (PGD) \citep{madry2018Deep} starts from a random point within the $x \pm th$ and perform $i$ gradient steps, to obtain the final adversarial sample $\hat{x}$.

\subsection{Common Corruptions}
We now consider a set of corruption functions $C$ such that $\mathbb{P}_C(c)$ approximate the real-world frequency of these corruptions.
We can now define, a classifier's robustness against corruptions as,
\begin{aequation}
\E_{c \sim C} [ \sP_{(x, y) \sim \train} ~~\underset{y}{\argmax} && f(c(x)) = y~~ ].
\end{aequation}
Thus, corruption robustness measures the classifier's average-case performance on classifier-agnostic corruptions $C$, while adversarial robustness measures the worst-case performance on small, additive, classifier-tailored perturbations.

\subsection{Adversarial Training}
The simplest and most straightforward way to defend against such adversarial attacks is to minimize the loss of adversarial examples, which is often called adversarial learning.
The adversarial learning framework proposed by \cite{madry2018Deep} does solve the following non-convex outer minimization problem and non-convex inner maximization problem, as follows:
\begin{aequation}
\underset{\theta}{\argmin} & & \mathbb{E}_{(x, y) \sim D} [~~ \underset{\Vert \epsilon_{x} \Vert_p ~\leq~ th}{\text{maximize}} & & \Ls_{\text{CE}, \theta}(f(\hat{x}^{i}), y) ~~ ].
\end{aequation}
There are various adversarial learning frameworks, including PGD \citep{madry2018Deep}, and TRADES \cite{zhang2019Theoretically} depending on how adversarial sample is optimized and how the classifier is optimized.

\subsection{Mixture of Expert Network}

A Mixture of Experts network consists of a set of $M$ `expert networks', $f_1, f_2, \ldots, f_M$ and a gating network which is generally set to be linear \citep{shazeer2017Outrageously, fedus2022Switch}.
The output of $m$-th expert network with input $x$ and parameter $W$ can be written as $f_m(x; W)$.
Similarly, the output of the gating network parameterised by
$\Theta = [ \theta_1, \ldots, \theta_m ] \in \mathbb{R}^{d \times M}$
can be written as
$h(x; \Theta) = \sum \Theta^{T} x$.
Thus, the output $F$ of the Mixture of Experts network can be written as follows,
\begin{aequation}
F(x; \Theta, W) = \sum_{m \in T_x} \pi_m(x; \Theta) f_m(x; W)
\end{aequation}
where $ T_x \subseteq [M] $ is a set of selected indices of experts and $ \pi_m(x; \Theta) $ is route gate value given by
\begin{aequation}
\pi_m(x; \Theta) = \frac{ \exp(h_m(x; \Theta))} { \sum_{m'=1}^{M} \exp(h_{m'}(x; \Theta))}, \quad \forall ~~ m \in [M]
\end{aequation}

\subsection{Mixture of Class-Specific Expert Network}

We define, a Mixture of Class-Specific Experts network consisting of a set of $N$ `expert networks', $f_1, f_2, \ldots, f_N$ and an aggregation network $F$.
The output of $n$-th expert network with input $x$ and parameter $\theta_n$ can be written as $f_{\theta_n}(x)$.
In our image classification problem, $f_{\theta_n}(x) \in \mathbb{R}^{2}$ is the probabilities (confidence) for either where the given $x$ belongs to $n$-th class or not.
Similarly, the output of aggregation network $F$ can be written as,
\begin{aequation}
F(x) &= \{ f_{\theta_n}(x)_{\text{True}} :  \forall ~~ n \in [N] \}
     &= \begin{pmatrix} f_{\theta_1}(x)_{\text{True}} \\  f_{\theta_2}(x)_{\text{True}} \\ \vdots \\ f_{\theta_N}(x)_{\text{True}} \end{pmatrix}
\end{aequation}
where, $f_{\theta_n}(x)_{\text{True}}$ is the true probability (confidence) of $n$-th expert that the sample $x$ belongs to $n$-th class.

The set of all parameters needed to be optimised can be written as $\Theta = {\theta_1, \ldots, \theta_n}$
Thus, the performance of Mixture of Class-Specific Expert Network on test queries drawn from data distribution $\train$ is,
\begin{aequation}
\mathbb{P}_{(x, y) \sim \train} ~~\underset{y}{\argmax} & & F(x) = y~~.
\end{aequation}

\section{Methodology}

\subsection{Common Corruptions}

The Common Corruptions benchmark \citep{hendrycks2019Benchmarking} introduced corrupted versions of commonly used classification datasets (ImageNet-C, CIFAR10-C) as standardized benchmarks which consists of $19$ diverse corruption types categoried into noise, blur, weather, and digital corruptions and each corruption type has five levels of severity.
This is because in real-world these corruptions can manifest themselves at varying intensities.
Since, the real-world corruptions also have variation even at a fixed intensity, to simulate these, the benchmark also introduces variation for each corruption when possible.
We  briefly describe each of the $19$ corruption in the benchmark below,

\begin{description}[itemsep=1.6pt, leftmargin=0pt]

  \item[Gaussian Noise,] this corruption can appear in low-lighting conditions.
  \item[Shot Noise,] also called Poisson noise, is electronic noise caused by the discrete nature of light itself.
  \item[Impulse Noise,] is a color analogue of salt-and-pepper noise and can be caused by bit errors.
  \item[Speckle Noise,] an additive noise where the noise added to a pixel tends to be larger if the original pixel intensity is larger.
  \item[Defocus Blur,] occurs when an image is out of focus.
  \item[Frosted Glass Blur,] appears with “frosted glass” windows or panels.
  \item[Motion Blur,] appears when a camera is moving quickly.
  \item[Zoom Blur,] occurs when a camera moves toward an object rapidly.
  \item[Gaussian Blur,] is a low-pass filter where a blurred pixel is a result of a weighted average of its neighbors, and farther pixels have decreasing weight in this average.
  \item[Snow,] is a visually obstructive form of precipitation.
  \item[Frost,] forms when lenses or windows are coated with ice crystals.
  \item[Fog,] shrouds objects and is rendered with the diamond-square algorithm.
  \item[Brightness,] varies with daylight intensity.
  \item[Spatter,] can occlude a lens in the form of rain or mud.
  \item[Contrast,] can be high or low depending on lighting conditions and the photographed object’s color.
  \item[Elastic Transformations,] stretch or contract small image regions.
  \item[Pixelation,] occurs when upsampling a lowresolution image.
  \item[JPEG Compression,] is a lossy image compression format which introduces compression artifacts.
  \item[Saturate,] is common in edited images where images are made more or less colorful.

\end{description}

\subsection{Evaluation Metric for Common Corruptions}

Common corruptions can be benign or destructive depending on their severity.
In order to comprehensively evaluate a classifier’s robustness to a given type of corruption, we score the classifier’s performance across five corruption severity levels and aggregate these scores.
The first evaluation step is to take a trained classifier $f$, and compute the performance on clean dataset ($P^{f}_{\text{clean}}$).
The second step is to test the classifier on each corruption type $c$ at each level of severity $s$ ($P^{f}_{c,s}$).

We then aggregate the classifier's performance for each of the $19$ corruption type $c$ as,
\begin{aequation}
\overline{P}^{f}_{c} = 1/5 \times \sum_{s=1}^{5} E_{f s,c}
\end{aequation}
Finally, we aggregate the classifier's average performance on the common corruptions as,
\begin{aequation}
\overline{P}^{f} = 1/19 \times \sum_{c} E_{f s,c}
\end{aequation}

\subsection{Mixture of Class-Specific Experts}

\begin{figure}[!t]
  \centering
  \includegraphics[width=\textwidth]{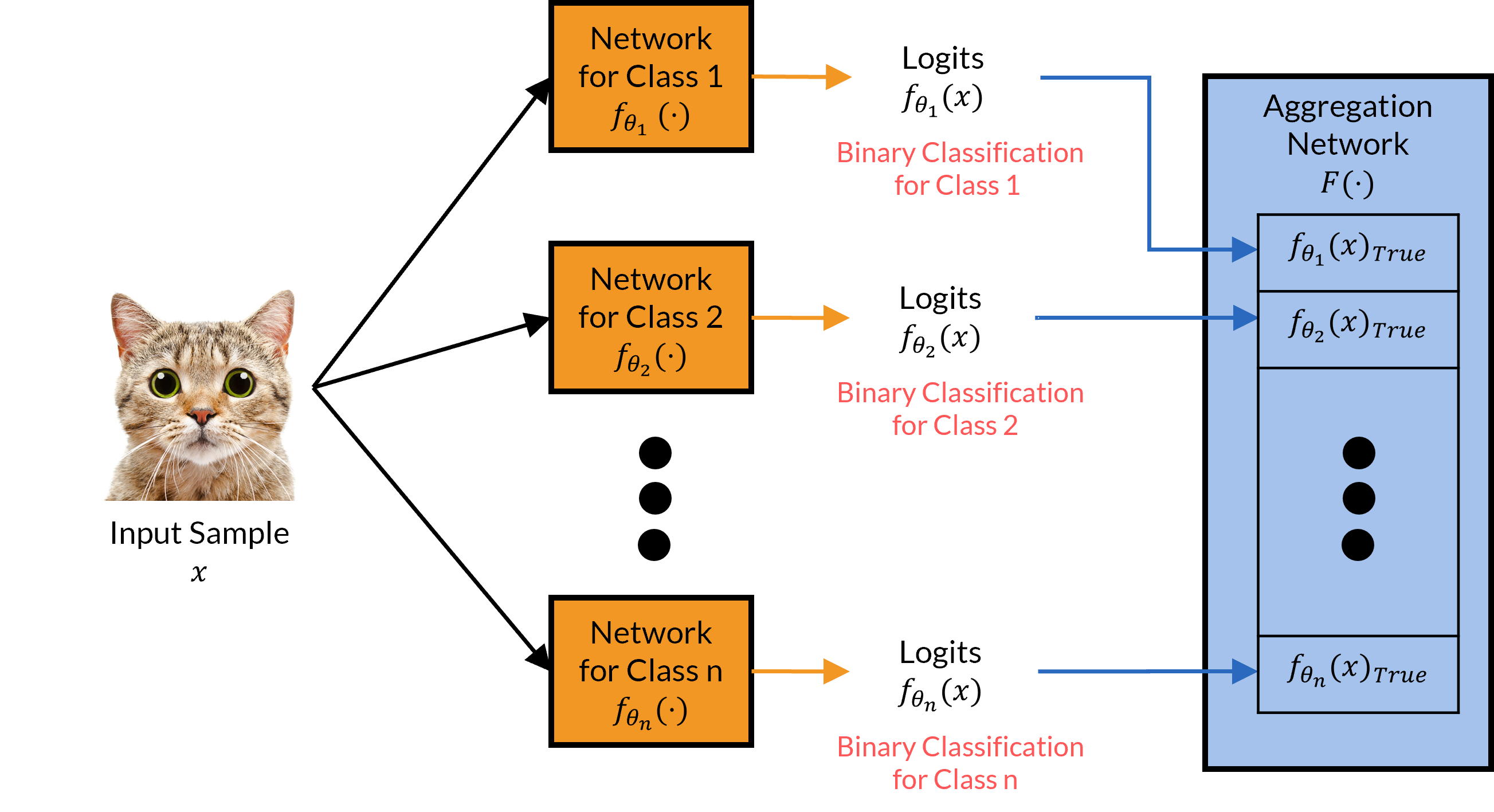}
  \caption{
    Illustration of the proposed Mixture of Class-Specific Expert Architecture.
  }
  \label{fig:mosce}
\end{figure}

\Figref{fig:mosce} provide an illustration for the proposed Mixture of Class Specific Expert Architecture.
We first create $N$ Class-Specific Expert Networks whose task will be to identify only if the given input sample is within their class and respond in binary logit representing whether the input sample belongs to the class or not.
We then create a simple non-parameterised aggregator network which takes the true soft-label from the Class-Specific Expert Networks and forms the final logit space of the entire model.
We then computed the cross-entropy loss $\mathcal{L}_{\text{CE}}$ using the aggregated logits, optimizing the parameters for all the class specific networks.
We also compute the binary cross-entropy loss $\mathcal{L}_{\text{BCE}}$ of individual Class-Specific Expert Network to optimise their own parameter.

\section{Results}

\begin{table}[!t]
  \centering
  \caption{Performance of Pre-activated ResNet-18 with different training strategies and dataset for image classification task. Performance is measured in Accuracy.}
  \resizebox{0.99\columnwidth}{!}{
  \begin{tabular}{l|l|rr|rr|rr}
    \toprule

    \multicolumn{2}{c|}{\textbf{Dataset}} & \multicolumn{4}{c|}{ \textbf{CIFAR-10}} &\multicolumn{2}{c}{ \textbf{CIFAR-10 Augmented}} \\
    \midrule
    \multicolumn{2}{c|}{\textbf{Training}} & \multicolumn{2}{c|}{\textbf{Standard}} & \multicolumn{2}{c|}{\textbf{Adversarial}} & \multicolumn{2}{c}{\textbf{Standard}} \\
    \midrule
    \multicolumn{2}{c|}{\textbf{Architecture}}& \textbf{Standard} & \textbf{MoCSE} & \textbf{Standard} & \textbf{MoCSE} & \textbf{Standard} & \textbf{MoCSE}\\
    \midrule
    \midrule
    \multicolumn{2}{l|}{\textbf{Natural Samples}} & \x{95.18\%} & 94.08\% & 83.35\% & \x{85.09\%} & 96.13\% & \x{96.54\%} \\
    \midrule
    \multirow{4}{*}{\textbf{Noise}}
    & \textbf{Gaussian Noise} & 17.63\% & \x{26.37\%} & 52.96\% & \x{59.83\%} & 16.69\% & \x{25.04\%} \\
    & \textbf{Shot Noise}     & 17.66\% & \x{25.66\%} & 50.51\% & \x{56.81\%} & 18.03\% & \x{24.67\%} \\
    & \textbf{Impulse Noise}  & 25.53\% & \x{33.17\%} & 52.06\% & \x{57.79\%} & 28.97\% & \x{37.48\%} \\
    & \textbf{Speckle Noise}  & 25.43\% & \x{35.77\%} & 60.12\% & \x{64.73\%} & 27.98\% & \x{34.75\%} \\
    \midrule
    \multirow{5}{*}{\textbf{Blur}}
    & \textbf{Defocus Blur}  & 16.68\% & \x{17.87\%} & \x{30.33\%} & 29.85\% & 14.20\% & \x{15.05\%} \\
    & \textbf{Glass Blur}    & 20.18\% & \x{23.71\%} & \x{39.64\%} & 38.56\% & 17.57\% & \x{22.81\%} \\
    & \textbf{Motion Blur}   & 30.69\% & \x{31.87\%} & \x{38.40\%} & 38.29\% & 30.73\% & \x{32.94\%} \\
    & \textbf{Zoom Blur}     & 71.48\% & \x{77.18\%} & 75.75\% & \x{76.61\%} & 76.13\% & \x{80.55\%} \\
    & \textbf{Gaussian Blur} & 19.74\% & \x{24.12\%} & \x{36.33\%} & 35.99\% & 17.74\% & \x{21.14\%} \\
    \midrule
    \multirow{5}{*}{\textbf{Weather}}
    & \textbf{Snow}       & \x{70.35\%} & 67.22\% & 51.37\% & \x{53.07\%} & 73.43\% & \x{75.03\%} \\
    & \textbf{Frost}      & \x{74.40\%} & 67.60\% & 40.73\% & \x{42.21\%} & 77.22\% & \x{79.15\%} \\
    & \textbf{Fog}        & \x{43.51\%} & 43.10\% & 17.41\% & \x{17.67\%} & 40.32\% & \x{44.43\%} \\
    & \textbf{Brightness} & \x{88.94\%} & 85.54\% & 64.11\% & \x{66.13\%} & 91.36\% & \x{92.03\%} \\
    & \textbf{Spatter}    & 82.67\% & \x{82.90\%} & 72.28\% & \x{75.55\%} & 85.29\% & \x{87.45\%} \\
    \midrule
    \multirow{5}{*}{\textbf{Digital}}
    & \textbf{Contrast}          & \x{52.08\%} & 41.88\% & \x{22.14\%} & 21.90\% & \x{52.43\%} & 49.78\% \\
    & \textbf{Elastic Transform} & 19.59\% & \x{19.70\%} & \x{40.98\%} & 40.95\% & 21.80\% & \x{24.03\%} \\
    & \textbf{Pixelate}          & 28.10\% & \x{36.73\%} & 66.55\% & \x{67.33\%} & \x{30.96\%} & 30.63\% \\
    & \textbf{JPEG Compression}  & 52.00\% & \x{55.87\%} & 76.08\% & \x{77.70\%} & 46.77\% & \x{51.02\%} \\
    & \textbf{Saturate}          & 86.08\% & \x{83.00\%} & 74.29\% & \x{75.78\%} & 89.22\% & \x{90.52\%} \\
    \midrule
    \rowcolor{Gray}
    \multicolumn{2}{c|}{\textbf{Average of 19 Corruptions}} & 44.35\% & \x{46.28\%} & 50.63\% & \x{52.46\%} & 45.10\% & \x{48.34\%} \\
    \bottomrule
  \end{tabular}
  }
  \label{tab:preact_resnet}
\end{table}

\textbf{Datasets.}
CIFAR-10 and CIFAR-10-Augmented

\textbf{Models.}
Standard Preact-ResNet-18 and Mixture of Class-Specific Experts Architecture

\textbf{Training Strategies.}
Standard Training and Adversarial Training

\textbf{Results.}
\Tableref{tab:preact_resnet} shows the result of image classifier, Preact-ResNet-18 in native form and in Mixture of Class-Specific Experts (MoCSE) form trained using standard procedure and using adversarial training recipe.
From the results, we can see that CIFAR-10-Augmented Dataset offers a boost in performance of ResNet by around $1\%$ for non-corrupted images and improve robustness of ViT by around $1\%$ for common corruptions.
We can see also see that Adversarial Training on the standard CIFAR-10 Dataset degrades the performance of the ResNet on non-corrupted images but improves robustness of ResNet in general by $8\%$ for common corruptions.
This results shows that Adversarial Training benefits in improving the robustness of the model, when the model capacity of the model is lower and a small dataset is used.
Further, augmented dataset help pushing the performance of the model even when the model is trained to the capacity suggesting that existing neural networks based on Convolution Layers can still benefit from Augmented Dataset.

When our Mixture of Class-Specific Experts Architecture is applied on ResNet,  we notice a general improvement in performance and robustness of ResNet, across all trainings such as Standard Training with CIFAR-10 Dataset, Standard Training with CIFAR-10 Augmented Dataset, and Adversarial Training with CIFAR-10 Dataset.
We improve the robustness of the ResNet trained with Standard Recipe using CIFAR-10 Dataset from $44.35\%$ to $46.28\%$.
We also improve the robustness of ResNet trained with Standard Recipe using CIFAR-10 Augmented Dataset from $45.10\%$ to $48.34\%$.
Further, we notice an improvement of best robustness of the ResNet trained with Adversarial Training Recipe using CIFAR-10 Dataset from $50.63\%$ to $52.46\%$.
This results show that Mixture of Class-Specific Experts Architecture enables better robustness of the model by disentangling the features across classes to learn such features which can better identify a sample belonging to a particular class.

\section{Conclusion}

In conclusion, the study acknowledges the inherent challenges in achieving neural network robustness to diverse image corruptions.
The proposed Mixture of Class-Specific Expert Architecture introduces a novel paradigm for enhancing robustness by disentangling feature learning, leading to improved performance across varied datasets.
As the pursuit of resilient models continues, the need for comprehensive evaluation methodologies, such as the common corruptions benchmark, becomes evident.
By addressing the limitations of current approaches and advancing our understanding of neural network vulnerabilities, this research aims to contribute to the development of more robust learning systems capable of adept reasoning in the face of real-world distortions.

\bibliography{iclr2024_conference}
\bibliographystyle{iclr2024_conference}


\end{document}